\documentclass[journal,twoside,web]{ieeecolor}
\usepackage{tmi}
\usepackage{cite}
\usepackage{amsmath,amssymb,amsfonts}
\usepackage{algorithmic}
\usepackage{graphicx}
\usepackage{textcomp}
\usepackage{dsfont}
\usepackage{color, soul}
\usepackage{diagbox}
\usepackage{makecell}

\def\BibTeX{{\rm B\kern-.05em{\sc i\kern-.025em b}\kern-.08em
    T\kern-.1667em\lower.7ex\hbox{E}\kern-.125emX}}
\usepackage{graphicx,amssymb,mathrsfs,amsmath,amsfonts,dsfont}
\usepackage{hyperref}
\usepackage{lineno}
\usepackage{stfloats}
\usepackage{algorithm}
\usepackage{algorithmic}
\usepackage{amsbsy}
\usepackage[mathscr]{euscript}
\usepackage{bm}
\usepackage{multirow}
\usepackage[utf8]{inputenc}
\newcommand{\w}{\mathbf{w}}
\newcommand{\x}{\mathbf{x}}
\newcommand{\s}{\mathbf{S}}
\newcommand{\y}{\mathbf{y}}
\newcommand{\z}{\mathbf{z}}
\newcommand{\D}{\mathrm{d}}


\begin{document}
\title{Physics-Informed DeepMRI: Bridging the Gap from Heat Diffusion to $k$-Space Interpolation}
\author{Zhuo-Xu~Cui, Congcong~Liu, Xiaohong~Fan, Chentao~Cao, Jing~Cheng,  Qingyong~Zhu, Yuanyuan~Liu, Sen~Jia, Yihang Zhou, Haifeng~Wang, Yanjie~Zhu, Jianping~Zhang, Qiegen~Liu, Dong~Liang   
       % <-this % stops a space
\thanks{This work was supported in part by the National Natural Science Foundation of China (62125111, 12026603, 62206273, 61771463, 81830056, U1805261, 81971611, 61871373, 81729003, 81901736).}
\thanks{Corresponding author: D. Liang (e-mail: dong.liang@siat.ac.cn)}
\thanks{Z.-X. Cui and C. Liu contributed equally to this work}
\thanks{Z.-X. Cui, Q. Zhu, and D. Liang are with Research Center for Medical AI, Shenzhen Institute of Advanced Technology, Chinese Academy of Sciences, Shenzhen, China.}
\thanks{X. Fan and J. Zhang are with School of mathematics and computational science, Xiangtan University, Xiangtan, China.}
\thanks{C. Liu, C. Cao, J. Cheng, S. Jia, Y. Zhou, H. Wang, Y. Zhu and D. Liang are with Paul C. Lauterbur Research Center for Biomedical Imaging, Shenzhen Institute of Advanced Technology, Chinese Academy of Sciences, Shenzhen, China.}
\thanks{Y. Liu is with National Innovation Center for Advanced Medical Devices, Shenzhen, China.}
\thanks{Q. Liu is with the Department of Electronic Information Engineering, Nanchang University, Nanchang, China}
\thanks{D. Liang is with Pazhou Lab, Guangzhou, China}
}

\maketitle

\begin{abstract}
In the field of parallel imaging (PI), alongside image-domain regularization methods, substantial research has been dedicated to exploring $k$-space interpolation. However, the interpretability of these methods remains an unresolved issue. Furthermore, these approaches currently face acceleration limitations that are comparable to those experienced by image-domain methods. In order to enhance interpretability and overcome the acceleration limitations, this paper introduces an interpretable framework that unifies both $k$-space interpolation techniques and image-domain methods, grounded in the physical principles of heat diffusion equations. Building upon this foundational framework, a novel $k$-space interpolation method is proposed. Specifically, we model the process of high-frequency information attenuation in $k$-space as a heat diffusion equation, while the effort to reconstruct high-frequency information from low-frequency regions can be conceptualized as a reverse heat equation. However, solving the reverse heat equation poses a challenging inverse problem. To tackle this challenge, we modify the heat equation to align with the principles of magnetic resonance PI physics and employ the score-based generative method to precisely execute the modified reverse heat diffusion.
Finally, experimental validation conducted on publicly available datasets demonstrates the superiority of the proposed approach over traditional $k$-space interpolation methods, deep learning-based $k$-space interpolation methods, and conventional diffusion models in terms of reconstruction accuracy, particularly in high-frequency regions.
\end{abstract}


\section{Introduction}
\label{sec:introduction}
\IEEEPARstart{M}{agnetic} resonance imaging (MRI) plays a pivotal role in routine clinical practice. However, the relatively slow pace of data acquisition has posed a longstanding challenge. The endeavor to reduce imaging time has become a central research objective. Consequently, there is a growing interest in reconstructing high-quality MR images from a limited amount of $k$-space data, aiming to accelerate the acquisition process \cite{liang1992constrained}.

Over the past two decades, substantial research efforts have been dedicated to image-domain parallel imaging (PI) methods \cite{sodickson1997simultaneous,Pruessmann1999SENSE}; however, these methods have shown certain limitations in reconstruction quality. In 2006, the highly influential concept of compressed sensing (CS) was introduced \cite{Candes2006Robust,Donoho2006Compressed,Lustig2007Sparse,cui2017nonconvex}. Drawing inspiration from CS, image-domain PI can be formulated as a sparse regularization framework \cite{Liang2009Accelerating,She2014Sparse,7163966,8017620}. Guided by the principles of CS theory, such methods exhibit robust interpretability and achieve high-quality reconstructions at specific acceleration rates. Nevertheless, as the demand for even higher acceleration rates continues to rise, these methods are no longer sufficient. Hence, there is an urgent need for the development of approaches capable of accommodating these higher acceleration tasks.

On the other hand, $k$-space PI methods, often considered distinct from image-domain PI, have also undergone significant development. $k$-space PI primarily relies on the ``predictable"  assumption, wherein missing data can be interpolated based on neighboring data \cite{Griswold2002Generalized,Lustig2010SPIRiT,8962389,10177777}. In comparison to image-domain PI, the interpretability of these methods remains an unresolved issue. Additionally, it's worth noting that empirical observations suggest that such approaches encounter similar acceleration limitations as those observed in image-domain methods.

Rethinking the image-domain and $k$-space PI methods, in the context of $k$-space PI, the typical approach involves acquiring low-frequency regions and estimating interpolation kernels to predict missing high-frequency data. By conceptualizing $k$-space PI as a procedure that anticipates high-frequency missing data based on low-frequency data, the inverse process implies a gradual attenuation of high-frequency information. This inherent mechanism shares a fundamental similarity with the principles underlying the heat diffusion equation. From a continuous standpoint, $k$-space linear interpolation methods can be seen as approximations of the reverse heat equation through estimated linear differential equations. Additionally, the gradient descent algorithm employed in the image-domain sparse regularization model can also be transmuted into the Perona-Malik (PM) equation \cite{56205} via continuousization, functioning as an approximation of the reverse heat equation. Consequently, both image-domain and $k$-space PI methods can be unified as approximations of the reverse heat equation. However, the attainment of the reverse heat equation involves solving an inverse problem (specifically, the first kind of Fredholm equation). The approximations introduced by image-domain (PM equation) and $k$-space (linear approximation) PI methods consequently lead to reduced accuracy. This constitutes the primary factor behind the limited acceleration rates observed in current image-domain and $k$-space PI methods.

\subsection{Contributions}
Building upon the aforementioned motivation, the primary objective of this study is to introduce an innovative PI technique that effectively approximates the reverse heat equation, enabling accurate reconstructions in scenarios characterized by high acceleration rates. To this end, the key contributions of this paper can be summarized as follows:

\begin{enumerate}
\item 
By employing the forward and reverse heat diffusion equations to model the process of attenuated high-frequency information and its reconstruction in $k$-space, this study introduces an innovative and interpretable framework for both $k$-space interpolation techniques and image-domain reconstruction methods.

\item Addressing the intricate nature of the reverse heat equation as an inverse problem, we tackle this challenge by adapting the heat equation to align with the principles of MR PI physics. Additionally, we introduce a score-based generative method to precisely execute the modified reverse heat diffusion, thereby achieving accurate reconstruction of missing high-frequency information.

\item Experimental validation conducted on publicly available datasets vividly showcases the superiority of the proposed approach when compared to traditional $k$-space interpolation methods, deep learning-based $k$-space interpolation techniques, and conventional diffusion models. The manifested improvements are evidenced in enhanced reconstruction quality, particularly in high-frequency regions.

\end{enumerate}

The remainder of the paper is organized as follows. Section \ref{sect2} describes the related works. Section \ref{sect3} discusses the methodology of the proposed method. The implementation details are presented in Section \ref{sect5}. Experiments performed on several datasets are presented in Section \ref{sect6}. A discussion is presented in Section \ref{sect7}. Section \ref{sect8} provides some concluding remarks.

\section{Related Work $\&$ Rethinking}\label{sect2}
Firstly, we present a summary of the mathematical notions and their corresponding notations discussed in the following sections, displayed in Table \ref{Symbol Meaning}.
\begin{table}
  \caption{\label{Symbol Meaning}Summary of mathematical notions and corresponding notations.}
  \centering
  \resizebox{0.85\linewidth}{!}{
      \begin{tabular}{c|c}
        \hline Notations & Notions \\
        \hline  $\mathbb{F}$ & Fourier transformation \\
        $\mathbf{x}$ & MR image \\
        $\mathbf{\hat{x}}$ & $k$-space data, $\mathbf{\hat{x}}=\mathbb{F}(\mathbf{x})$ \\
        $\mathbf{M}$ & undersampling operator \\
        $\mathbf{S}$ & coil sensitivity maps, $\mathbf{S}=[\mathbf{s}^*_1, \mathbf{s}^*_2, ..., \mathbf{s}^*_m]^*$, $\mathbf{S^*S}=\mathbf{I}$ \\
        $\mathbf{\bar{S}}$ & $\mathbf{\bar{S}}:=\mathbb{F}\mathbf{S}\mathbb{F}^{-1}$\\
        $\mathbf{A}$ & encoding matrix, $\mathbf{A}=\mathbf{M}\mathbb{F}\mathbf{S}$ \\
        $\mathbf{y}$ & undersampled $k$-space data, $\mathbf{y}=\mathbf{Ax}$\\
        $\mathbf{G}_t$ & Gaussian function \\
        $\dot{\mathbf{G}}_t$ &  derivative of $\mathbf{G_t}$ with respect to $t$, $\dot{\mathbf{G}}_t=\D\mathbf{G_t}/\D t $\\
        $\mathbf{I}$ & identity operator \\
        $\nabla$ & gradient\\
        $\nabla\cdot$ &divergence\\
        $\Delta$ & Laplace operator, $\Delta=\nabla\cdot\nabla$ \\
        $\odot$ & element-wise multiplication  \\
        $\circledast$ &convolution\\
        \hline 
    \end{tabular}
    }
\end{table}
\subsection{PI methods}
\subsubsection{Image-Domain PI methods}
In the context of parallel acquisition, an image-domain PI model can be formulated into a redundant linear equation, i.e., 
$$\y=\mathbf{A}\x$$
where $\y$ represents the acquired multi-channel under-sampled $k$-space data, $\x$ is the desired MR image, and $\mathbf{A}$ stands for the MR signal multi-channel encoding system.
The original SENSE algorithm aims to solve the above equations to reconstruct an MR image \cite{pruessmann1999sense1}. With the advent of CS, the SENSE model has been reformulated to a sparsity-regularized form \cite{Knoll2021Parallel}, i.e.,
$$\x^*=\arg\min_{\x}\frac{1}{2}\|\y-\mathbf{A}\x\|^2+\lambda \|\nabla \x\|_1$$
Moreover, applying the gradient descent algorithm to the above model leads to the PM equation, combined with data consistency through continuousization:
\begin{equation}
\label{pm}\frac{\D \x}{\D t}=-\mathbf{A}^*(\mathbf{A}\x-\y)-\lambda \nabla\cdot\left(\frac{\nabla \x}{\|\nabla \x\|_1}\right). 
\end{equation}
In simpler terms, in the image domain, the forward heat equation can be understood as a process of image blurring, while the PM equation is designed for image deblurring. In essence, the PM equation serves as a manually designed approximation of the reverse heat equation.

\subsubsection{$k$-Space PI methods}
Alternatively, a PI model can be formulated in $k$-space as an interpolation procedure, assuming that the values of $k$-space data within each channel are predictable within a neighborhood. Prominent examples of $k$-space PI models include GRAPPA \cite{Griswold2002Generalized}, SPIRiT \cite{Lustig2010SPIRiT}, etc.
It is worth noting that, by introducing the concept of limits, extends the interpolation kernel estimation of GRAPPA to learning mappings from low to high frequencies (the GRAPPA operator \cite{griswold2005parallel}). Specifically, let $\widehat{\x}(\w)$ denote the $i$th column of the $k$-space (representing low-frequency information), and $\widehat{\x}(\w+\Delta_{\w})$ denote the $i+1$th column (representing high-frequency information). The GRAPPA operator maps $\widehat{\x}(\w)$ to $\widehat{\x}(\w+\Delta_{\w})$, thus estimating high-frequency information missing from the low-frequency estimation. If we abstract $\widehat{\x}(\w)$ and $\widehat{\x}(\w+\Delta_{\w})$ as $k$-space signals at different evolution time, that is, $\widehat{\z}(t):=\widehat{\x}(\w)$ and $\widehat{\z}(t+\Delta_t):=\widehat{\x}(\w+\Delta_w)$, following the concept of the GRAPPA operator, there exists an operator $\mathbf{K}_{\Delta_t}$ such that $\widehat{\z}(t+\Delta_t)=\mathbf{K}_{\Delta_t}\widehat{\z}(t)$ and $\lim_{\Delta_t\to 0} \mathbf{K}_{\Delta_t}=\mathbf{I}$. 
Consequently, the GRAPPA operator can lead to a time-evolution equation from low to high frequencies, given by 
\begin{equation}\label{grappa}
{\D\widehat{\z}}=\mathbf{P}\widehat{\z}{\D t}
\end{equation}
where $\mathbf{P}:=\lim_{\Delta_t\to 0}(\mathbf{K}_{\Delta_t}-\mathbf{I})/\Delta_t$.
In the subsequent discussion, we will model the gradual attenuation of high-frequency information in $k$-space as a heat equation. Therefore, the GRAPPA operator can be seen as an approximation of the reverse heat equation through the linear equation (\ref{grappa}).

Nevertheless, obtaining the reverse heat equation requires solving an inverse problem, specifically the first kind of Fredholm equation. Consequently, accurately approximating the reverse heat equation using both (\ref{pm}) and (\ref{grappa}) becomes a challenging task. This compels us to redefine the procedure of high-frequency information attenuation within the framework of MR PI principles and to develop algorithms that can faithfully execute its reverse, thus enabling the accurate reconstruction of missing high-frequency information.
\subsection{Score-Based Diffusion Model}
The score-based diffusion model serves as a framework for diffusion generative models \cite{NEURIPS2020_4c5bcfec,NEURIPS2019_3001ef25,score-based-SDE}. It introduces incremental Gaussian noise at various scales to perturb data, progressively molding the data distribution into a Gaussian form. Subsequently, it generates samples from Gaussian noise based on the corresponding reverse procedure. Specifically, the diffusion process $\{\mathbf{x}(t)\}_{t=0}^T$ can be seen as the solution of the forward SDE as follows:
\begin{equation}
    \mathrm{d} \mathbf{x}=\mathbf{f}(\mathbf{x}, t) \mathrm{d} t+\mathbf{V}(t) \mathrm{d} \mathbf{w},
    \label{background: forward sde}
\end{equation}
where $t$ is the continuous time variable, $t \in [0, T]$, $\mathbf{x}(0) \sim p_0=p_{data}$, $\mathbf{x}(T) \sim p_T$ and $p_T$ is a prior distribution, typically using Gaussian distribution. $\mathbf{f}$ and $g$  are the drift and diffusion coefficients of $\mathbf{x}(t)$, and $\mathbf{w}$ is the standard Wiener process. The reverse-time SDE of (\ref{background: forward sde}) is:
\begin{equation}
    \mathrm{d} \mathbf{x}=\left[\mathbf{f}(\mathbf{x}, t)-\mathbf{V}(t)\mathbf{V}(t)^* \nabla_{\mathbf{x}} \log p_{t}(\mathbf{x})\right] \mathrm{d} t+\mathbf{V}(t) \mathrm{d} \mathbf{\bar w},
\end{equation}
where $\mathbf{\bar w}$ is the standard Wiener process for the time from $T$ to $0$. The score function $\nabla_{\mathbf{x}} \log p_{t}(\mathbf{x})$ is approximated by the score model $\mathbf{s}_{\boldsymbol{\theta}}$ trained by
\begin{multline}
        \boldsymbol{\theta}^{*}=\underset{\boldsymbol{\theta}}{\arg \min } \mathbb{E}_{t}\Big\{\lambda(t) \mathbb{E}_{\mathbf{x}(0)} \mathbb{E}_{\mathbf{x}(t) \mid \mathbf{x}(0)}\big[\big\|\mathbf{s}_{\boldsymbol{\theta}}(\mathbf{x}(t), t)\\-\nabla_{\mathbf{x}(t)} \log p_{0 t}(\mathbf{x}(t) \mid \mathbf{x}(0))\big\|_{2}^{2}\big]\Big\},
    \label{SDE-loss}
\end{multline}
where $p_{0 t}(\mathbf{x}(t) \mid \mathbf{x}(0))$ is the perturbation kernel and can be derived from the forward diffusion process. Once the score model $\mathbf{s}_{\boldsymbol{\theta}}$ is trained, we can generate samples through reverse-time SDE.

It is worth noting that \cite{rissanen2022generative} addressed the case where 
$\mathbf{f}(\mathbf{x}, t)$ in equation (\ref{background: forward sde}) corresponds to a Laplace operator and $\mathbf{V}(t)=0$, thereby causing (\ref{background: forward sde}) to reduce into a heat diffusion process. Meanwhile, \cite{daras2022soft} explores the scenario where 
$\mathbf{f}(\mathbf{x}, t)$ represents an arbitrary linear operator, effectively extending (\ref{background: forward sde}) to a general diffusion process, including heat diffusion. However, this paper distinguishes itself from these methodologies in two crucial aspects:
\begin{enumerate}
\item Firstly, it introduces a novel approach by conceptualizing the attenuation of high-frequency $k$-space information as a heat diffusion process—a facet that had not been previously explored in their investigation.
\item Secondly, in the upcoming sections, we will further refine the heat diffusion process to align it more closely with the fundamental principles of MR PI physics. Moreover, through our ablation experiments, we will compare our approach with the methodologies without refinements. This comparative analysis will highlight the crucial significance of validating the refinements, which are grounded in the principles of MR PI physics.
\end{enumerate}

\section{Methodology}\label{sect3}
In this section, we will present the underlying principle of modeling $k$-space attenuation using heat diffusion. Following that, we will refine the heat diffusion model to align with the fundamental principles of MR PI physics and execute its inverse process by employing a score-based generative model.

\subsection{Modeling $k$-Space Attenuation Through Heat Diffusion} 

\begin{figure}[thbp]
\centerline{\includegraphics[width=0.48\textwidth]{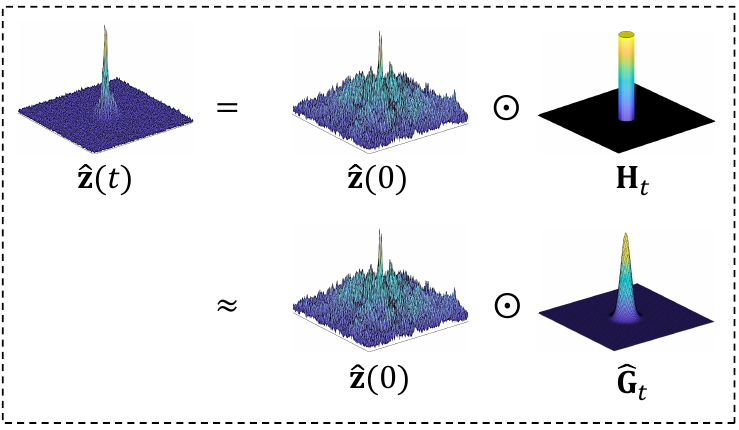}}
	\caption{Illustration of $k$-space attenuation. $\widehat{\z}(0)$ represents the fully sampled $k$-space data, while $\widehat{\z}(0)$ denotes the high-frequency missing $k$-space data.}
	\label{f1}
\end{figure}

Inspired by the GRAPPA operator, as time evolves, equation (\ref{grappa}) generates high-frequency information from the low-frequency domain, which can be regarded as the inverse process of $k$-space high-frequency attenuation. Intuitively, $k$-space attenuation can be achieved through the following procedure: 
\begin{equation}\label{attenuation:1}\widehat{\z}(t)=\mathbf{H}_t\odot\widehat{\z}(0)\end{equation}
where $\mathbf{H}_t$ is an indicator function satisfying:
\begin{equation*}
\mathbf{H}_t(\w):= \begin{cases}c, & |\w| \leqslant r(t) \\ 0, & \text { otherwise }\end{cases}
\end{equation*}
and $r(t)$ is a function that decreases with time $t$. Since the indicator function can be approximated by a Fourier transformed Gaussian function, thus, equation (\ref{attenuation:1}) can be approximated as:
\begin{equation}\label{attenuation:2}\widehat{\z}(t)=\mathbf{\widehat{G}}_t\odot\widehat{\z}(0)\end{equation}
where $\mathbf{\widehat{G}}_t=\mathbb{F}(\mathbf{G}_t)$ satisfying $\lim_{t \to 0}\mathbf{\widehat{G}}_t=\mathbf{I}$ and  $\lim_{t \to +\infty}\mathbf{\widehat{G}}_t=\delta$. The $k$-space attenuation process is vividly depicted in Figure \ref{f1}.
According to the convolution theorem, performing inverse Fourier transforms on both sides of equation (\ref{attenuation:2}) results in $$\z(t)=\mathbf{G}_t\circledast\z(0)$$ which represents the solution to the heat equation 
\begin{equation*}
{\D\z}=\Delta\z{\D t}.
\end{equation*}
Hence, it becomes evident that the $k$-space attenuation process can be delineated by the heat equation, while image-domain PI (\ref{pm}) and $k$-space PI (\ref{grappa}) can both be regarded as approximations of the reverse heat equation.

However, acquiring the reverse heat equation involves solving the first kind of Fredholm equation, which inherently possesses ill-posed characteristics and is intricate to solve with high precision. Consequently, both equations (\ref{pm}) and (\ref{grappa}) do not provide precise approximations of the reverse heat equation. This fundamental challenge serves as the primary factor limiting the achievable acceleration rates in the existing image-domain and $k$-space PI methods.
\subsection{Attenuated $k$-Space Diffusion}
Due to the limitations imposed by the ill-posed nature of the reverse heat equation, we will refine the $k$-space attenuation model (heat equation) based on the principles of MR PI physics, aiming to facilitate the solvability of its reverse process.
\subsubsection{Forward SDE} Firstly, we transform the heat equation into the $k$-space, we derive
$$
\D \widehat{\z}=\dot{\mathbf{\widehat{G}}}(t) \odot \widehat{\z}(0) \D t 
$$
Considering the context of MR signal acquisition, Gaussian noise is commonly encountered. Additionally, due to the multi-coil acquisition, the Gaussian noise must still conform to the distribution pattern of coil sensitivities. Consequently, we introduce a noise term into the aforementioned equation, resulting in  
\begin{equation}\label{Forward-SDE}
  \D \widehat{\z}=\dot{\mathbf{\widehat{G}}}_t \odot \widehat{\z}(0) \D t+\sqrt{\frac{\D \sigma(t)^2}{\D t}} \bar{\s}\bar{\s}^* \D \mathbf{w}  
\end{equation}
where $\sigma(t)$ is the parameter to control the noise level, $\bar{\s}$ represents $\mathbb{F}\mathbf{S}\mathbb{F}^{-1}$, $\s$ denotes coil sensitivity, and $\bar{\s}\bar{\s}^*$ is introduced to ensure the added noise conforms to the distribution pattern of coil sensitivities.

\subsubsection{Reverse SDE}
\begin{figure*}[!t]
    \centerline{\includegraphics[width=1\textwidth]{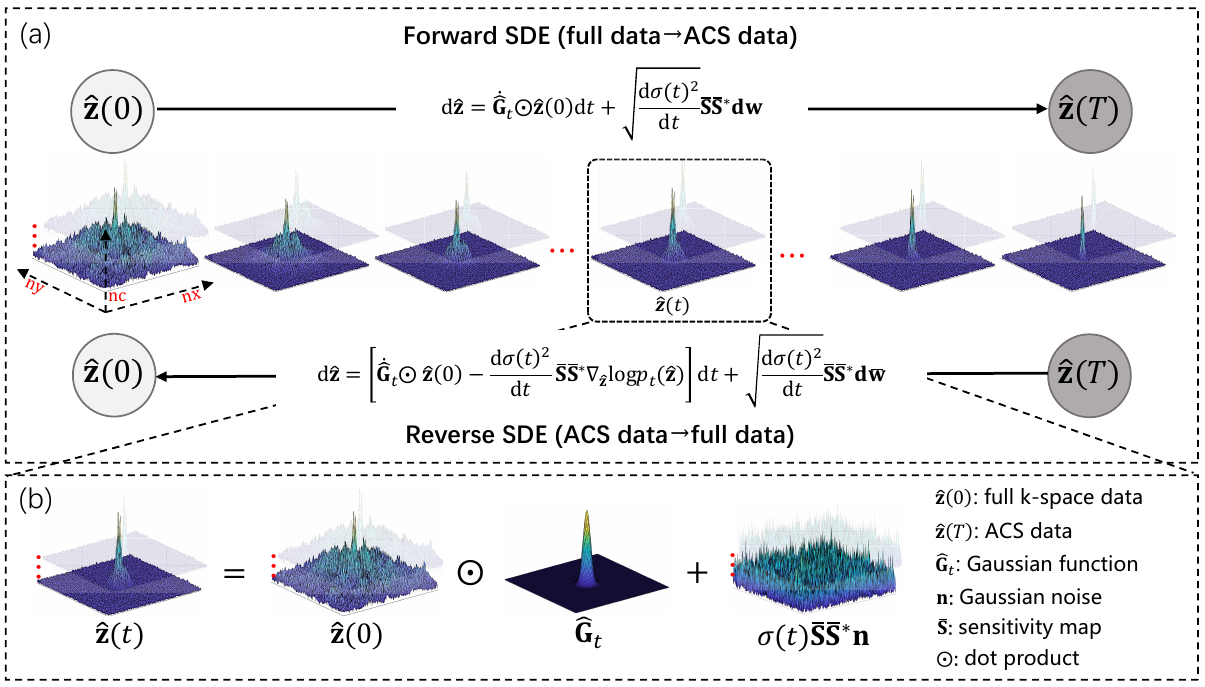}}
    \caption{The framework of attenuated $k$-space diffusion. (a) In the forward process, the fully sampled $k$-space data undergoes heat diffusion to transform into low-frequency ACS data, while noise conforming to the coil sensitivity distribution is gradually incorporated into the multi-channel $k$-space data. In the reverse process, high-frequency information is gradually reconstructed based on the noised low-frequency ACS data. (b) In the forward process $t$: the fully sampled $k$-space data is element-wise multiplied by a Gaussian function and noise, consistent with the coil sensitivity distribution, is added to obtain $\widehat{\z}(t)$.}
    \label{fig:general}
\end{figure*}

By introducing noise term, we have modeled $k$-space attenuation as SDE (\ref{Forward-SDE}). In contrast to the heat equation, according to the theory of SDE \cite{anderson1982reverse}, there exists a reverse SDE for equation (\ref{Forward-SDE}), enabling the completion of missing $k$-space data. In particular, the reverse SDE of (\ref{Forward-SDE}) reads:
\begin{equation}\label{Reverse-SDE}
    \begin{aligned}
\D \widehat{\z}=&\left[\dot{\mathbf{\widehat{G}}}_t \odot \widehat{\z}(0)-\frac{\D \sigma(t)^2}{\D t} \bar{\s} \bar{\s}^* \nabla_{\widehat{\z}} \log p_t(\widehat{\z})\right] \D t\\&+\sqrt{\frac{\D \sigma(t)^2}{\D t}}\bar{\s} \bar{\s}^* \D \bar{\mathbf{w}}
    \end{aligned}
\end{equation}
Reverse SDE (\ref{Reverse-SDE}) involves an unknown function $\nabla_{\widehat{\z}} \log p_t(\widehat{\z})$. Next, we will elucidate how to learn $\nabla_{\widehat{\z}} \log p_t(\widehat{\z})$ through the score-matching method \cite{vincent2011connection}. The whole framework of attenuated $k$-space diffusion is shown in Figure \ref{fig:general}

\subsubsection{Estimating Score Functions}
To estimate $\nabla_{\widehat{\z}} \log p_t(\widehat{\z})$ using the score-matching method, it is essential to obtain the perturbation kernel of SDE (\ref{Reverse-SDE}). According to Eqs. 5.50 and 5.51 in \cite{sarkka2019applied}, the perturbation kernel is given by:
\begin{equation*}
    p_{0 t}(\widehat{\z}(t)\mid \widehat{\z}(0)) = \mathcal{N}\big(\widehat{\z}(t); \mathbf{\widehat{G}}_t\odot\widehat{\z}(0), [\sigma^2(t)-\sigma^2(0)] \bar{\s}\bar{\s}^*\big).
    \label{perturbation kernel}
\end{equation*}
According to score-matching method (\ref{SDE-loss}), $\nabla_{\widehat{\z}} \log p_t(\widehat{\z})$ can be obtained by solving the following optimization problem 
\begin{multline*}
\boldsymbol{\theta}^{*}=\underset{\boldsymbol{\theta}}{\arg \min } \mathbb{E}_{t}\Big\{\lambda(t) \mathbb{E}_{\widehat{\z}(0)} \mathbb{E}_{\widehat{\z}(t) \mid \widehat{\z}(0)}\big[\big\|\mathbf{s}_{\boldsymbol{\theta}}(\widehat{\z}(t), t)\\+\frac{\mathbf{\widehat{G}}_t\odot\widehat{\z}(0)-\widehat{\z}(t)}{\sigma^2(t)\bar{\s}\bar{\s}^*} \big\|_{2}^{2}\big]\Big\},
\end{multline*}
and estimated from $\mathbf{s}_{\boldsymbol{\theta^*}}$.
Let
\begin{equation}\label{prior}\mathbf{s}_{\boldsymbol{\theta}}(\widehat{\z}(t), t):=\frac{\mathbf{\widehat{G}}_t\odot\mathbf{h}_{\boldsymbol{\theta}}(\widehat{\z}(t), t)-\widehat{\z}(t)}{\sigma^2(t)\bar{\s}\bar{\s}^*} \end{equation}
and multiply the above optimization objective by $\bar{\s}^*\bar{\s}\bar{\s}^*$. As a result, the score-matching loss function is reduced to:
\begin{equation}\label{loss}\begin{aligned}
\boldsymbol{\theta}^*=\underset{\boldsymbol{\theta}}{\arg \min }& \mathbb{E}_t\left\{\lambda(t) \mathbb{E}_{\widehat{\mathbf{z}}(0)} \mathbb{E}_{\widehat{\mathbf{z}}(t) \mid \widehat{\mathbf{z}}(0)}\left[\right.\right.\\
&\left.\left.\left\|\bar{\mathbf{S}}^*\left(\widehat{\mathbf{G}}_t \odot\left(\mathbf{h}_{\boldsymbol{\theta}}(\widehat{\mathbf{z}}(t), t)-\widehat{\mathbf{z}}(0)\right)\right)\right\|_2^2\right]\right\}
\end{aligned}\end{equation}
In particular, following the approach outlined in reference \cite{daras2022soft}, the network $\mathbf{h}_{\boldsymbol{\theta}}(\widehat{\mathbf{z}}(t), t)$ adopts a residual structure, namely, $\mathbf{h}_{\boldsymbol{\theta}}(\widehat{\mathbf{z}}(t), t)=\mathbf{r}_{\boldsymbol{\theta}}(\widehat{\mathbf{z}}(t), t)+\widehat{\mathbf{z}}(t)$.
\subsubsection{$k$-Space Interpolation Algorithm}
Based on the modeling of forward and reverse attenuated $k$-space diffusion, as well as the estimation of the prior term (\ref{prior}), performing discrete equation (\ref{Reverse-SDE}), i.e.,  
\begin{equation*}\begin{aligned}\widehat{\z}_i=&\widehat{\z}_{i+1}-(\mathbf{\widehat{G}}_{i+1}-\mathbf{\widehat{G}}_{i})\odot{\widehat{\z}}_0+\sqrt{\sigma_{i+1}^2-\sigma_{i}^2}\bar{\s}\bar{\s}^*\mathbf{n}\\
&+(\sigma_{i+1}^2-\sigma_{i}^2)\bar{\s}\bar{\s}^*\nabla_{\widehat{\z}_{i+1}} \log p_{i+1}(\widehat{\z}_{i+1}),\end{aligned}\end{equation*}
enables the reconstruction of missing high-frequency $k$-space data using low-frequency ACS data. It's worth noting that during the iterative process, ${\widehat{\z}}_0$ is not directly accessible. However, drawing from the loss function (\ref{loss}), the trained network $\mathbf{h}_{\boldsymbol{\theta}^*}({\widehat{\z}}_i,i)$ can be interpreted as a projection from ${\widehat{\z}}_i$ to ${\widehat{\z}}_0$. Nonetheless, in practical scenarios, relying solely on the network projection might not yield accurate results. In such cases, inspired by \cite{tu2022wkgm,chung2023fast}, we can correct the network projection using a $k$-space PI model. A typical $k$-space PI model is expressed as
\begin{equation*}
\min_{\hat{\mathbf{z}}} \|\mathcal{H}(\widehat{\z})\mathbf{N}\|_F^2~ \text { s.t. } \mathbf{M}\widehat{\z}=\y
\end{equation*}
where $\|\mathcal{H}(\widehat{\z})\mathbf{N}\|_F^2$ represents the structural low-rank (SLR) term, $\mathcal{H}$ usually represents Hankelization, $\mathbf{N}$ represents the annihilation filter and $\mathbf{M}\widehat{\z}=\y$ denotes data consistency \cite{8962381,cui2023k}. Within the vicinity of network projection, we will seek a solution for the aforementioned PI model, which will be utilized as the corrected ${\widehat{\z}}_0$. In particular, the above process is coupled with the Predictor-Corrector method (PC Sampling), which is in detail illustrated in Algorithm \ref{alg:1}.
\begin{algorithm}[htb]
	\caption{PC Sampling (Attenuated $k$-Space Diffusion).}
	\label{alg:1}
	\begin{algorithmic}[1]
		\STATE {\bfseries Input:} $\{\mathbf{\widehat{G}}_i\}_{i=1}^N, \{\sigma_i\}_{i=1}^N, \mathbf{M}, \bar{\mathbf{S}}, \mathbf{y}, \lambda, r, N, M$;\\
		\STATE {\bfseries Initialize:} $\widehat{\z}_{N} \sim \mathcal{N}(\mathbf{\widehat{G}}_N\odot\y, \sigma^2_N \bar{\s}\bar{\s}^*)$;\\
		\FOR{$i = N-1$ to $0$}
            \STATE $\mathbf{n} \sim \mathcal{N}(\mathbf{0}, \mathbf{I})$;
		\STATE $ {\widehat{\z}'}_0\leftarrow\mathbf{h}_{\boldsymbol{\theta}^*}(\widehat{\mathbf{z}}_{i+1}, i+1)$;
           \STATE ${\widehat{\z}''}_0\leftarrow\arg\min_{\widehat{\z}}\frac{1}{2}\|\mathbf{M}\widehat{\z}-\y\|^2+\|\mathcal{H}(\widehat{\z})\mathbf{N}\|_F^2+\lambda\|\widehat{\z}-\widehat{\z}'\|^2$;
           \STATE $\bm{\epsilon}_{i+1}\leftarrow \frac{\mathbf{\widehat{G}}_{i+1}\odot\mathbf{h}_{\boldsymbol{\theta}}(\widehat{\z}_{i+1}, i+1)-\widehat{\z}_{i+1}}{\sigma^2_{i+1}\bar{\s}\bar{\s}^*}$;
           \STATE $\widehat{\z}_i\leftarrow\widehat{\z}_{i+1}-(\mathbf{\widehat{G}}_{i+1}-\mathbf{\widehat{G}}_{i})\odot{\widehat{\z}''}_0+(\sigma_{i+1}^2-\sigma_{i}^2)\bar{\s}\bar{\s}^*\bm{\epsilon}_{i+1}+\sqrt{\sigma_{i+1}^2-\sigma_{i}^2}\bar{\s}\bar{\s}^*\mathbf{n} $;
		\FOR{$k=1$ to $M$}
             \STATE $\mathbf{n} \sim \mathcal{N}(\mathbf{0}, \mathbf{I})$;
		\STATE $\mathbf{g} \leftarrow\frac{\mathbf{\widehat{G}}_{i}\odot\mathbf{h}_{\boldsymbol{\theta}^*}(\widehat{\z}_{i}, i)-\widehat{\z}_{i}}{\sigma^2_{i}\bar{\s}\bar{\s}^*}$;
        \STATE $\eta \leftarrow 2 \left(r\|\mathbf{n}\|_{2} /\|\mathbf{g}\|_{2}\right)^{2}$;
        \STATE $\widehat{\z}_{i} \leftarrow \widehat{\z}_{i}+\eta\bar{\s}\bar{\s}^*\mathbf{g}+\sqrt{2 \eta} \bar{\s}\bar{\s}^*\mathbf{n}$;
		\ENDFOR
		\ENDFOR
		\STATE {\bfseries Output:} $\widehat{\z}_{0}.$\\
	\end{algorithmic}
\end{algorithm}

\section{Implementation}\label{sect5}

\subsection{Data Acquisition}
 The FastMRI knee raw data \footnote{\url{https://fastmri.org/}} was acquired from a 3T Siemens scanner (Siemens Magnetom Skyra, Prisma and Biograph mMR). Data acquisition used a 15 channel knee coil array and conventional Cartesian 2D TSE protocol employed clinically at NYU School of Medicine. The following sequence parameters
were used: Echo train length 4, matrix size $320 \times 320$, in-plane resolution $0.5mm\times0.5mm$, slice thickness $3mm$, no gap between slices. Timing varied between systems, with repetition time (TR) ranging between 2200 and 3000 milliseconds, and echo time (TE) between 27 and 34 milliseconds. From them, we randomly select T1-weighted data of 34 individuals (1002 slices in total) as the training set and data of 3 individuals (95 slices in total) as the test set.

\subsection{Network Architecture and Training}
The network structure of attenuated $k$-space diffusion is the same as that of VE-diffusion (\texttt{ncsnpp}\footnote{\url{https://github.com/yang-song/score_sde_pytorch}\label{sde code}}). The exponential moving average (EMA) rate is set to $0.999$, the number of iterations $N$ and $M$ is set to 50 and 1, respectively, $\sigma_{N}=1$, $\sigma_{0}=0.01$ (we will explore the effects of different values of $\sigma$ on the reconstruction results in the Discussion section), and the batch size is set to 1. Unlike previous approaches that combine multi-coil data into a single channel for network training, our methods directly input multi-coil $k$-space data to the network. The complex $k$-space data is split into real and imaginary components and concatenated before input into the network, resulting in an input tensor of size $nc\times 2 \times nx \times ny$. $nc$ is the coil number, $2$ represents the concatenated real and imaginary parts of the data, and $nx$ and $ny$ represent the image size. The coil dimension is permuted to the batch size dimension to keep the convolution parameters of each channel consistent. The network is trained for 100 epochs in a computing environment using the torch1.13 library\cite{paszke2019pytorch}, cuda11.6 on an NVIDIA A800 Tensor Core GPU.

\subsection{Performance Evaluation}
In this study, the quantitative evaluations were all calculated on the image domain. The image is derived using an inverse Fourier transform followed by an elementwise square-root of sum-of-the-squares (sos) operation. For quantitative evaluation, the peak signal-to-noise ratio (PSNR), normalized mean square error (NMSE) value and structural similarity (SSIM) index \cite{1284395} were adopted.

\section{Experimentation Results }\label{sect6}
\subsection{Ablation Studies}
Differing from conventional diffusion models, the model introduced in this paper incorporates two distinct operations. Firstly, the noise introduced into the diffusion model aligns consistently with the distribution of coil sensitivities. Secondly, within the iterative process, an SLR $k$-space PI model is integrated to rectify the generated results. In this section, we will conduct ablation experiments to confirm the efficacy of these two operations, respectively.

\begin{figure}[thbp]
\centerline{\includegraphics[width=0.48\textwidth]{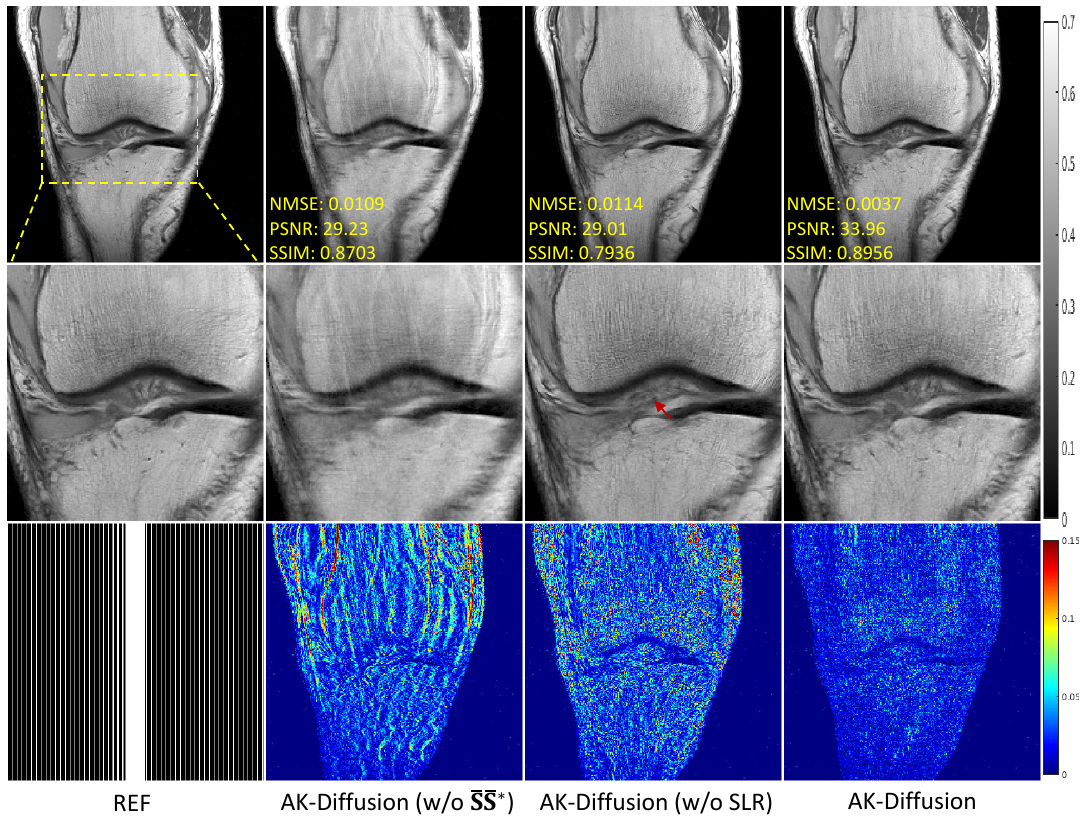}}
	\caption{Reconstruction results under uniform undersampling at $R=6$. The values in the corner are each slice's NMSE/PSNR/SSIM values. The second and third rows illustrate the enlarged and error views, respectively. The grayscale of the reconstructed images and the error images' color bar are on the figure's right.}
	\label{f2}
\end{figure}

Initially, we verified the effectiveness of maintaining a consistent distribution between the added noise and coil sensitivities. To achieve this, we design an ablation approach wherein we transform the proposed attenuated $k$-space diffusion (referred to as AK-Diffusion) into isotropic diffusion. Specifically, we substitute the $\bar{\s}\bar{\s}^*$ operator in the forward process (\ref{Forward-SDE}), reverse process (\ref{Reverse-SDE}), and loss function (\ref{loss}) with the identity operator $\mathbf{I}$, resulting in what we term as AK-Diffusion (w/o $\bar{\s}\bar{\s}^*$). Figure \ref{f2} showcases the reconstruction outcomes of AK-Diffusion both with and without $\bar{\s}\bar{\s}^*$ under uniform undersampling by a factor of 6. It is apparent that omitting $\bar{\s}\bar{\s}^*$ in AK-Diffusion substantially compromises the quality of the reconstruction results. The quantitative metrics in Table \ref{tab:1} correspondingly validate the performance consistency with visual perception. Thus, the pivotal role of $\bar{\s}\bar{\s}^*$ in the proposed AK-Diffusion is evident.

\begin{table}
	\begin{center}
		\caption{Quantitative comparison for ablation studies on the fastMRI knee dataset.}\label{tab:1}
		\setlength{\tabcolsep}{0.7mm}{
  \scriptsize
			\begin{tabular}{l|l|ccc}
				\hline
				\multicolumn{ 2}{c}{ Datasets} & \multicolumn{ 3}{|c}{Quantitative Evaluation}  \\
				\multicolumn{ 2}{c|}{ \& Methods   } &NMSE &PSNR(dB)&SSIM  \\
				\hline
				\multirow{3}{*}{\makecell{Reconstruction \\(uniform 6x)}}
				& AK-Diffusion (w/o $\bar{\s}\bar{\s}^*$) &0.0236$\pm$0.0190&28.47$\pm$2.85&0.87$\pm$0.03 \\
				\cline{2-5}
				& AK-Diffusion (w/o SLR)  &0.0135$\pm$0.0061&30.14$\pm$1.70&0.82$\pm$0.05 \\
				\cline{2-5}
				&  AK-Diffusion  &\textcolor{red}{0.0024$\pm$0.0043}&\textcolor{red}{34.90$\pm$2.04}& \textcolor{red}{0.90$\pm$0.04}\\
				\hline
		\end{tabular}}
	\end{center}
\end{table}

Next, we develop an ablation strategy to validate the significance of the coupled SLR model. For this purpose, in the sixth line of Algorithm \ref{alg:1}, we eliminate the SLR term $\|\mathcal{H}(\widehat{\z})\mathbf{N}\|_F^2$ and retain the data consistency term $\|\mathbf{M}\widehat{\z}-\y\|^2$, establishing the ablation approach referred to as AK-Diffusion (w/o SLR). Figure \ref{f2} further depicts the reconstruction outcomes of AK-Diffusion with and without SLR regularization under uniform undersampling by a factor of 6. The red arrow in Figure \ref{f2} marks the region where the removal of SLR correction is evident, leading to noticeable distortion in the reconstructed image details. This ablation experiment effectively validates the efficacy of SLR correction.

\subsection{Comparative Studies}
To demonstrate the effectiveness of the proposed method, a series of extensive comparative experiments were conducted in this section. 
Specifically, we compared to traditional $k$-space PI method, GRAPPA operator \cite{griswold2005parallel}, and structural low-rank model, AC-LORAKS \cite{haldar2013low}.
To validate the advantages of the diffusion model, we will compare it with an end-to-end $k$-space interpolation deep learning method, referred to as H-DSLR \cite{pramanik2020deep}. Additionally, to assess the benefits of the proposed diffusion equations (\ref{Forward-SDE}) and (\ref{Reverse-SDE}), we will conduct a comparison with the image-domain VE-diffusion model \cite{score-based-SDE}. In particular, for a fair comparison, VE-diffusion incorporated the PI correction as our proposed algorithm, following the methodology outlined in \cite{chung2023fast}.

\begin{figure*}[!t]
    \centerline{\includegraphics[width=1\textwidth]{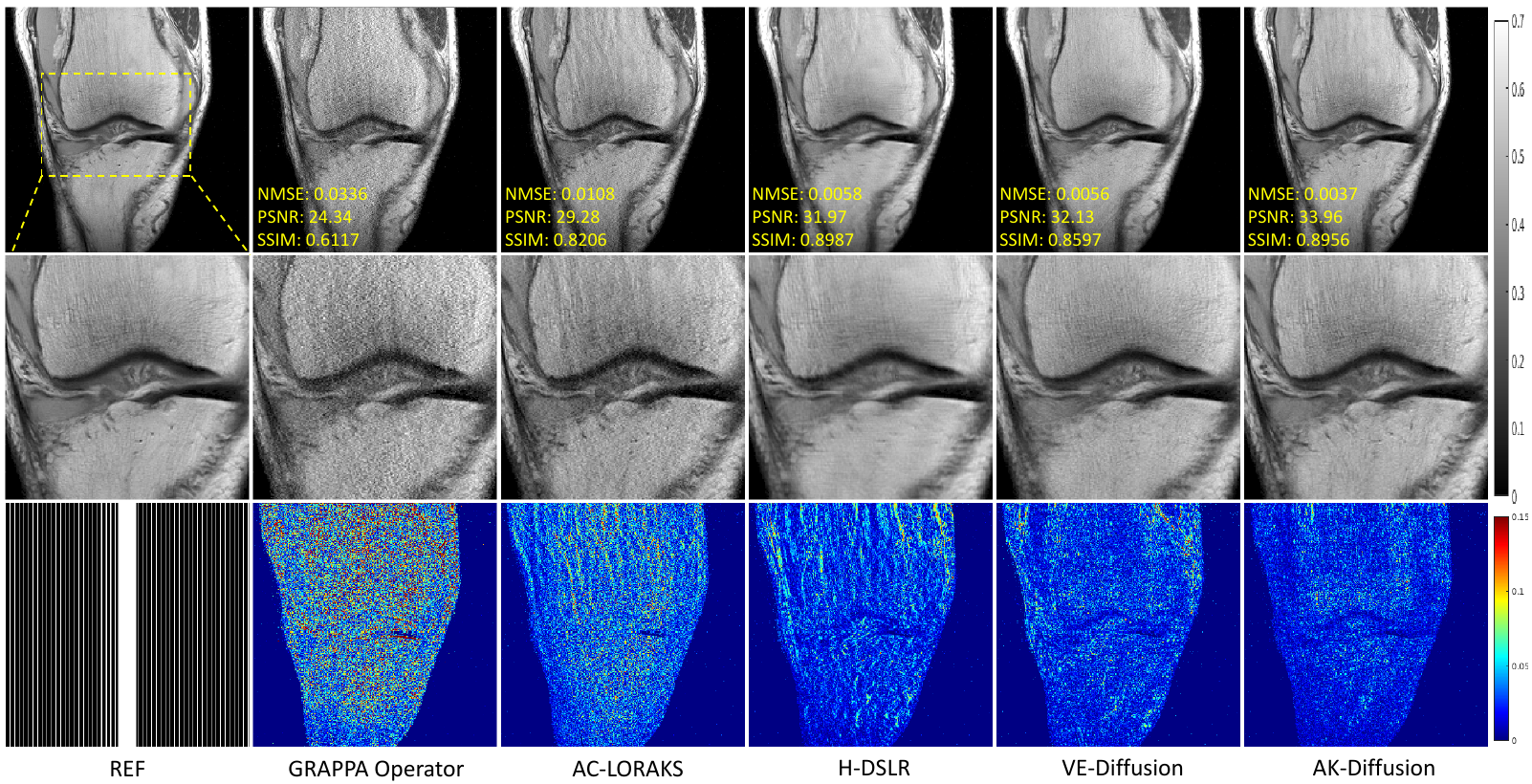}}
    \caption{Reconstruction results under uniform undersampling at $R=6$. The values in the corner are each slice's NMSE/PSNR/SSIM values. The second and third rows illustrate the enlarged and error views, respectively. The grayscale of the reconstructed images and the error images' color bar are on the figure's right.}
    \label{f3}
\end{figure*}

Figure \ref{f3} presents the reconstruction outcomes of the various methods under a uniform undersampling factor of 6. The results clearly indicate that the GRAPPA operator, AC-LORAKS, and H-DSLR yield aliasing patterns in their reconstructions. While VE-Diffusion successfully suppresses aliasing patterns, upon closer examination in the enlarged view, it becomes evident that compared to our AK-Diffusion, it sacrifices high-frequency details in its reconstructions. Table \ref{tab:2} complements these visual observations with quantitative metrics, further confirming the effectiveness of our proposed approach.

\begin{figure}[thbp]
\centerline{\includegraphics[width=0.48\textwidth]{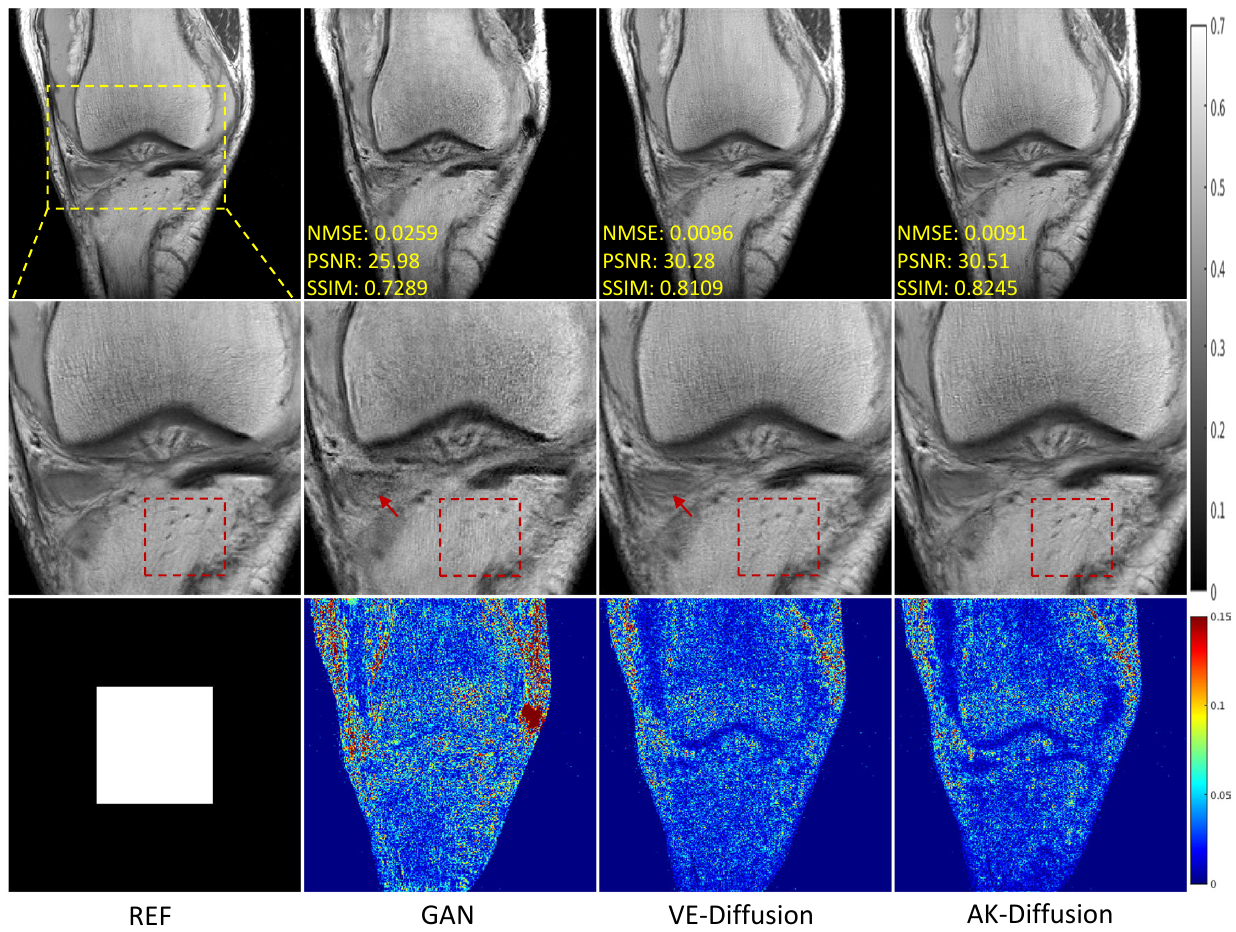}}
	\caption{Super-Resolution results under $128\times 128$ ACS region. The values in the corner are each slice's NMSE/PSNR/SSIM values. The second and third rows illustrate the enlarged and error views, respectively. The grayscale of the reconstructed images and the error images' color bar are on the figure's right.}
	\label{f4}
\end{figure}

To evaluate our model's performance in generating high-frequency data from low-frequency data, we conducted a super-resolution experiment. Specifically, we employed an undersampling pattern that included only a $128\times 128$ ACS region. We compared our method with GAN \cite{NIPS2014_5ca3e9b1} and VE-diffusion. The super-resolution results obtained using different methods are illustrated in Figure \ref{f4}. While GAN exhibits prominent artifacts, VE-diffusion performs better but still retains artifacts, as indicated by the red arrow. Additionally, in terms of high-frequency detail reconstruction, highlighted by the red-boxed region, our proposed AK-diffusion achieves the most accurate reconstruction. These experiments collectively affirm the accuracy of our method in generating high-frequency data.
\\
\\
\\

\begin{table}[h]
	\begin{center}
		\caption{Quantitative comparison for various methods on fastMRI knee dataset.}\label{tab:2}
		\setlength{\tabcolsep}{1.5mm}{
  \scriptsize
			\begin{tabular}{c|l|cccc}
				\hline
				\multicolumn{ 2}{c}{ Datasets} & \multicolumn{ 3}{|c}{Quantitative Evaluation}  \\
				\multicolumn{ 2}{c|}{ \& Methods   } &NMSE &PSNR(dB)&SSIM   \\
				\hline
				\multirow{5}{*}{\makecell{Reconstruction \\(uniform 6x)}}
				&  GRAPPA Op &0.0346$\pm$0.0128&25.96$\pm$2.25& 0.63$\pm$0.08\\
				\cline{2-5}
    				& AC-LORAKS &{0.0105$\pm$0.0040}&{31.18$\pm$2.23}& {0.81$\pm$0.05}\\
				\cline{2-5}
				&  H-DSLR &{0.0109$\pm$0.0096}&{31.85$\pm$2.27}&{0.90$\pm$0.03}\\
    				\cline{2-5}
    			& VE-Diffusion &{0.0106$\pm$0.0076}&{31.53$\pm$2.28}& 0.86$\pm$0.05\\
                        \cline{2-5}
				&  AK-Diffusion  &\textcolor{red}{0.0024$\pm$0.0043}&\textcolor{red}{34.90$\pm$2.04}& 0.90$\pm$0.04\\
    				\hline
				\multirow{3}{*}{\makecell{Super-Resolution \\(128x128 ACS)}}
    				& GAN &{0.0325$\pm$0.0151}&{26.39$\pm$1.80}& {0.74$\pm$0.08}\\
				\cline{2-5}
				&  VE-Diffusion &{0.0106$\pm$0.0053}&{31.25$\pm$2.17}&{0.82$\pm$0.07}\\
    				\cline{2-5}
    			& AK-Diffusion &\textcolor{red}{0.0099$\pm$0.0052}&\textcolor{red}{31.61$\pm$2.40}& \textcolor{red}{0.84$\pm$0.07}\\
				\hline
		\end{tabular}}
	\end{center}
\end{table}

\section{Discussion}\label{sect7}
In this paper, we introduced a forward AK-diffusion model to represent the attenuation process of $k$-space data. Subsequently, we employed a score-based generative method to ensure precise execution of the reverse AK-diffusion, enabling $k$-space interpolation. Through comprehensive comparative experiments, we substantiated the advantages of our proposed method in uniform undersampling reconstruction and super-resolution tasks. Ablation experiments further verified the roles of SLR PI correction and the incorporation of noise consistent with coil sensitivity distribution within our model. However, several aspects of our method warrant further discussion.

\subsection{Performance under Other Undersampling Patterns}
\begin{figure*}[!t]
    \centerline{\includegraphics[width=1\textwidth]{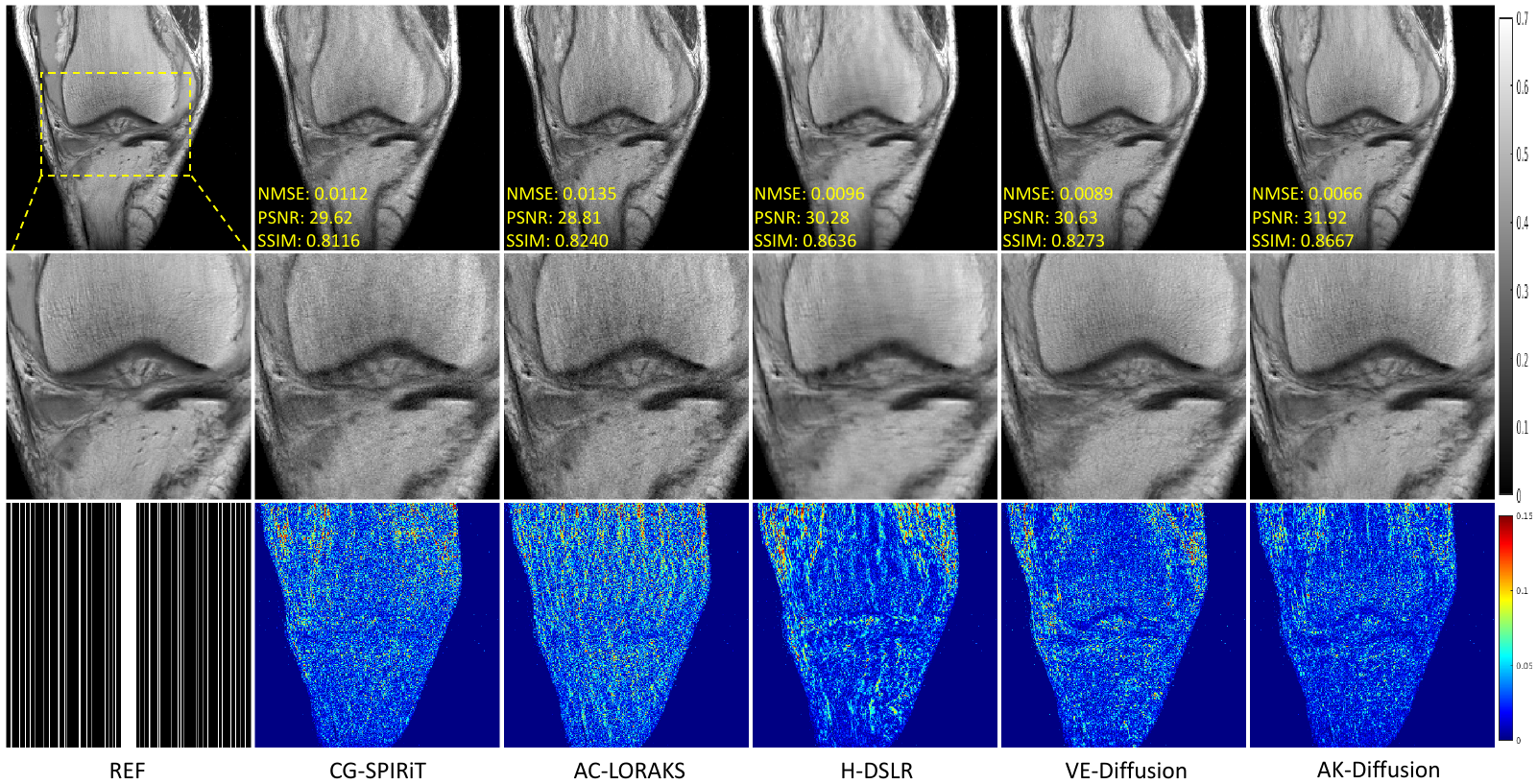}}
    \caption{Reconstruction results under random undersampling at $R=6$. The values in the corner are each slice's NMSE/PSNR/SSIM values. The second and third rows illustrate the enlarged and error views, respectively. The grayscale of the reconstructed images and the error images' color bar are on the figure's right.}
    \label{f5}
\end{figure*}

The forward and reverse diffusion processes, as well as the SLR PI correction integrated into our proposed AK-diffusion, are not confined to specific undersampling patterns. Consequently, our approach can be adapted to address other undersampled reconstruction scenarios. To affirm this, we present the reconstruction outcomes of various methods under random undersampling with a factor of 6 in Figure \ref{f5}. The results illustrate that CG-SPIRiT, AC-LORAKS, and H-DSLR exhibit aliasing patterns in their reconstructions. While VE-diffusion effectively suppresses aliasing patterns, its reconstructions display a notable loss of high-frequency details upon closer examination. In contrast, our proposed method not only effectively mitigates aliasing but also excels in preserving intricate image details. Table \ref{tab:3} provides quantitative metrics that align with visual observations, thus confirming the superior performance of our approach in comparative experiments involving random undersampling reconstruction.

\begin{table}[h]
	\begin{center}
		\caption{Quantitative comparison for various methods on fastMRI knee dataset.}\label{tab:3}
		\setlength{\tabcolsep}{0.4mm}{
  \scriptsize
			\begin{tabular}{c|l|cccc}
				\hline
				\multicolumn{ 2}{c}{ Datasets} & \multicolumn{ 3}{|c}{Quantitative Evaluation}  \\
				\multicolumn{ 2}{c|}{ \& Methods   } &NMSE &PSNR(dB)&SSIM   \\
				\hline
				\multirow{5}{*}{\makecell{Reconstruction \\(random 6x)}}
				&  CG-SPIRiT &0.0160$\pm$0.0072&29.37$\pm$2.15& 0.77$\pm$0.06\\
				\cline{2-5}
    				& AC-LORAKS &{0.0129$\pm$0.0046}&{30.22$\pm$2.00}& {0.78$\pm$0.05}\\
				\cline{2-5}
				&  H-DSLR &{0.0098$\pm$0.0036}&{31.46$\pm$1.14}&{0.86$\pm$0.04}\\
    				\cline{2-5}
    			& VE-Diffusion &{0.0091$\pm$0.0031}&{31.72$\pm$1.49}& 0.82$\pm$0.06\\
                        \cline{2-5}
				&  AK-Diffusion  &\textcolor{red}{0.0087$\pm$0.0045}&\textcolor{red}{32.13$\pm$2.00}& 0.86$\pm$0.05\\
    				\hline
				\multirow{6}{*}{\makecell{Reconstruction \\(uniform 6x)}}
    				& AK-Diffusion (sos)&{0.0053$\pm$0.0024}&{34.23$\pm$1.99}& {0.89$\pm$0.04}\\
				\cline{2-5}
				&  AK-Diffusion (SLR)&{0.0058$\pm$0.0026}&{33.83$\pm$1.86}&{0.89$\pm$0.04}\\
    				\cline{2-5}
    			& AK-Diffusion (ESPIRiT)&\textcolor{red}{0.0045$\pm$0.0024}&\textcolor{red}{34.90$\pm$2.04}& \textcolor{red}{0.90$\pm$0.04}\\
				\cline{2-5}
    				& AK-Diffusion ($\sigma_N=0.25$)&{0.0051$\pm$0.0024}&{34.33$\pm$1.97}& {0.89$\pm$0.04}\\
				\cline{2-5}
				&  AK-Diffusion ($\sigma_N=0.5$)&{0.0050$\pm$0.0023}&{34.49$\pm$2.01}&{0.89$\pm$0.04}\\
    				\cline{2-5}
    			& AK-Diffusion ($\sigma_N=1$)&\textcolor{red}{0.0045$\pm$0.0024}&\textcolor{red}{34.90$\pm$2.04}& \textcolor{red}{0.90$\pm$0.04}\\
				\hline
		\end{tabular}}
	\end{center}
\end{table}

\subsection{Robustness to $\bar{\s}\bar{\s}^*$}
\begin{figure}[thbp]
\centerline{\includegraphics[width=0.48\textwidth]{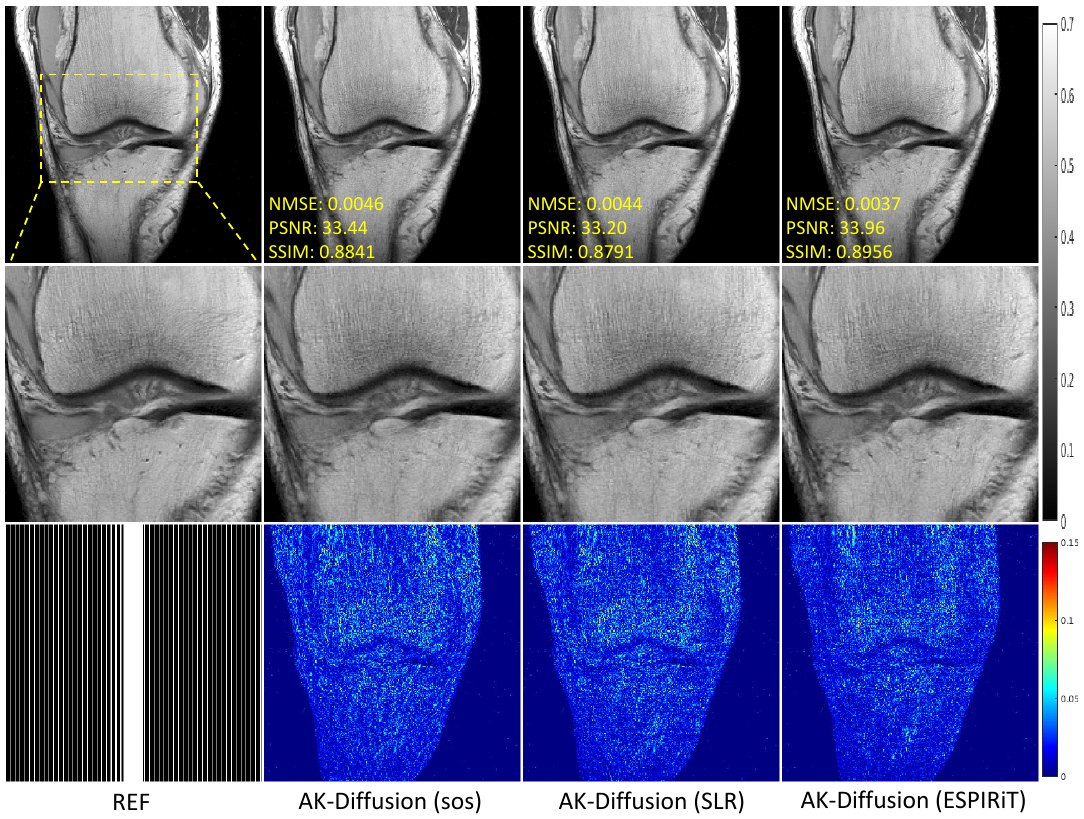}}
	\caption{Reconstruction results under uniform undersampling at $R=6$. The values in the corner are each slice's NMSE/PSNR/SSIM values. The second and third rows illustrate the enlarged and error views, respectively. The grayscale of the reconstructed images and the error images' color bar are on the figure's right.}
	\label{f6}
\end{figure}

Within our proposed method, the $\bar{\s}\bar{\s}^*$ operator, related to coil sensitivity estimation, often introduces additional computational complexity compared to conventional $k$-space interpolation methods. The ablation experiments have already underscored the crucial role of $\bar{\s}\bar{\s}^*$. We now assess our model's robustness to variations in $\bar{\s}\bar{\s}^*$ by evaluating whether its performance degrades when substituting computationally simpler yet less accurate coil sensitivity estimates. While we employed coil sensitivity estimated by ESPIRiT for model training and the aforementioned experiments, denoted as AK-diffusion (ESPIRiT), we also consider coil sensitivity estimation through the division of the multi-channel zero-filled image by its sos. This approach, termed AK-diffusion (sos), is evaluated during testing. Additionally, following ALOHA \cite{lee2016acceleration}, the structural low-rankness can characterize coil redundancy. Therefore, we can define the operator $\mathcal{T}(\hat{\mathbf{z}}):=\arg\min_{\hat{\mathbf{z}}} \|\mathcal{H}(\widehat{\z})\mathbf{N}\|_F^2$ as an alternative to $\bar{\s}\bar{\s}^*$ in AK-diffusion, referred to as AK-diffusion (SLR). Figure \ref{f6} demonstrates the performance of AK-diffusion under a uniform undersampling factor of 6 using these three different estimation approaches. The error view reveals that AK-diffusion with the ESPIRiT estimation method consistently employed during training achieves optimal performance. However, in terms of the visual perceptual quality of the reconstructed images, the performances of the three approaches are closely matched. Quantitative metrics in Table \ref{tab:3} further support this observation, emphasizing that while $\bar{\s}\bar{\s}^*$ is pivotal, our AK-diffusion model exhibits robustness to variations in it.

\subsection{Impact of the Noise Scale $\sigma_N$}
\begin{figure}[thbp]
\centerline{\includegraphics[width=0.48\textwidth]{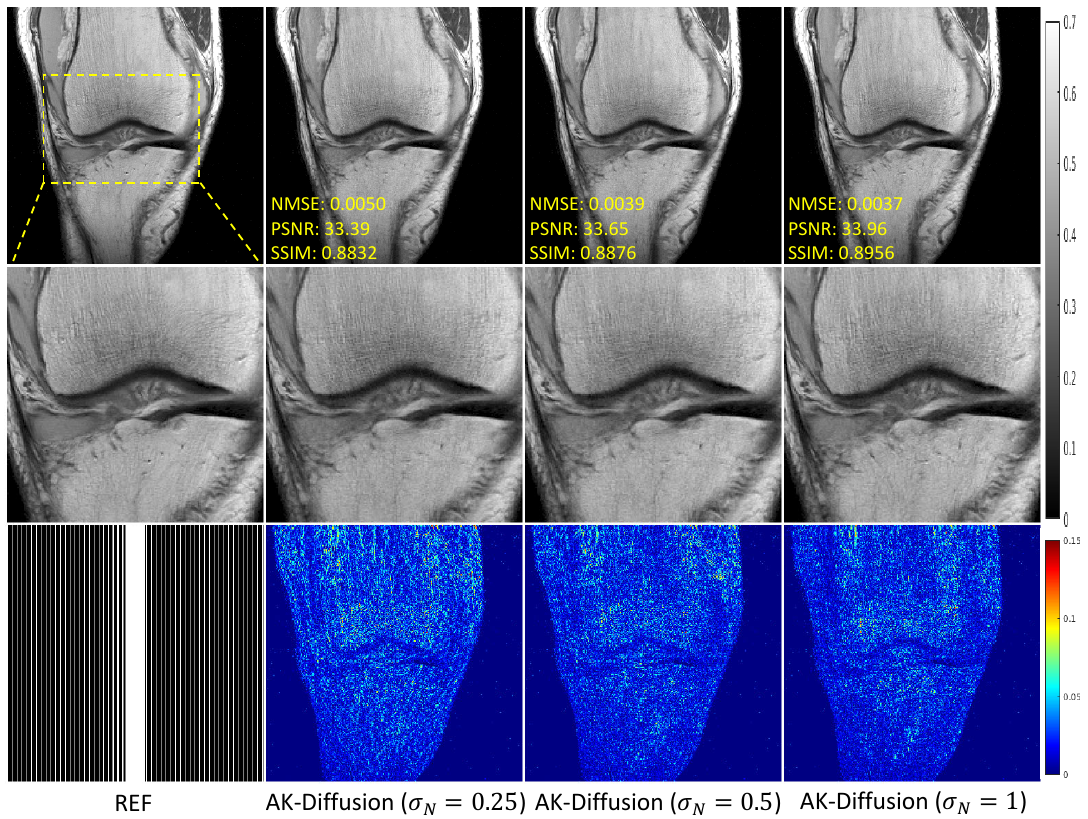}}
	\caption{Reconstruction results under uniform undersampling at $R=6$. The values in the corner are each slice's NMSE/PSNR/SSIM values. The second and third rows illustrate the enlarged and error views, respectively. The grayscale of the reconstructed images and the error images' color bar are on the figure's right.}
	\label{f7}
\end{figure}

Reflection on the design of AK-diffusion (\ref{Forward-SDE}) and (\ref{Reverse-SDE}) indicates that the introduction of the noise term transforms the heat equation into an SDE, ensuring the existence of its reverse process. However, the selection of the noise scale lacks theoretical guarantees. In our previous experiments, we empirically opted for $\sigma_N=1$. To investigate the noise scale's impact on reconstruction outcomes, we conducted additional comparative experiments with $\sigma_N=0.25$ and 0.5. Figure \ref{f7} illustrates the reconstruction outcomes of AK-diffusion under uniform undersampling with a factor of 6 for these different noise scales. Analysis of the error view and quantitative metrics in Table \ref{tab:3} reveals that increasing the noise scale positively influences reconstruction quality. However, closer observation highlights that with increased noise scale, high-frequency details in the reconstructed images become more prominent, potentially leading to the generation of pseudo-details. Therefore, the choice of $\sigma_N=1$ represents an empirical trade-off.

\section{Conclusion}\label{sect8}
In this paper, we have established a model that portrays the attenuation process of $k$-space data as an analogy to a heat diffusion. Furthermore, recognizing the inherent difficulty of solving the reverse heat diffusion equation, we presented a unified explanation for the acceleration limitations of both $k$-space and image-domain PI methods.
To address the intricacies posed by the reverse heat diffusion equation, we have refined the heat equation to align with the underlying principles of MR PI physics. Additionally, we have employed a score-based generative approach to execute the refined reverse heat diffusion process. Lastly, through comprehensive experimentation involving accelerated imaging and super-resolution tasks on publicly accessible datasets, we have substantiated the merits of our proposed method in terms of reconstruction precision, particularly in high-frequency domains.

\bibliographystyle{ieeetr}
\bibliography{library_manu}

\end{document}